\documentclass[10pt,twocolumn,letterpaper]{article}

\usepackage[pagenumbers]{cvpr} 

\makeatletter
\@namedef{ver@everyshi.sty}{}
\makeatother
\usepackage{tikz}

\usepackage{graphicx}
\usepackage{amsmath}
\usepackage{amssymb}
\usepackage{booktabs}

\usepackage{xcolor}
\usepackage{algorithm}
\usepackage{listings}
\usepackage{mathtools}
\usepackage{algpseudocode}
\usepackage{makecell} 
\usepackage{multirow}
\usepackage{pifont}
\usepackage{enumitem}
\usepackage{arydshln}
\usepackage{multirow}
\usepackage{xcolor}
\usepackage{wrapfig}
\usepackage{caption}
\usepackage[most]{tcolorbox}
\usepackage{subcaption}
\usepackage{float}
\newcommand{\br}[1]{\left({#1}\right)}

\usepackage[pagebackref=false, breaklinks=true, colorlinks,
            citecolor=citecolor, linkcolor=linkcolor, bookmarks=false]{hyperref}

\definecolor{citecolor}{HTML}{0071BC}
\definecolor{linkcolor}{HTML}{ED1C24}
\hypersetup{                    
  colorlinks,
  linkcolor={green!80!black},
  urlcolor={blue!90!black}
}

\definecolor{codeblue}{rgb}{0.25, 0.5, 0.5}
\definecolor{codekw}{rgb}{0.35, 0.35, 0.75}
\lstdefinestyle{Pytorch}{
    language         = Python,
    backgroundcolor  = \color{white},
    basicstyle       = \ttfamily\footnotesize,
    columns          = fullflexible,
    breaklines       = true,
    captionpos       = b,
    commentstyle     = \fontsize{4pt}{4pt}\color{codeblue},
    keywordstyle     = \fontsize{4pt}{4pt}\color{codekw},
    morekeywords     = with,
}

\newcommand{\cmark}{\ding{51}}%
\newcommand{\xmark}{\ding{55}}%
\newcommand{\ms}[1]{\textcolor{black}{#1}}

\makeatletter\renewcommand\paragraph{\@startsection{paragraph}{4}{\z@}{.5em \@plus1ex \@minus.2ex}{-.5em}{\normalfont\normalsize\bfseries}}\makeatother

\usepackage[capitalize]{cleveref}
\crefname{section}{Sec.}{Secs.}
\Crefname{section}{Section}{Sections}
\Crefname{table}{Table}{Tables}
\crefname{table}{Tab.}{Tabs.}

\def\cvprPaperID{***} 
\def\confName{CVPR}
\def\confYear{2023}

\newcommand{\authorskip}{\hspace{4.5mm}}

\begin{document}

\title{
Single Image Backdoor Inversion via Robust Smoothed Classifiers
}

\author{\quad\, Mingjie Sun$^{1}$\authorskip\,  J. Zico Kolter$^{1,2}$\\
[2.mm]
$^1$Carnegie Mellon University \quad $^2$Bosch Center for AI\\
}
\maketitle
\begin{abstract}
\ms{Backdoor inversion, a central step in many backdoor defenses, is a reverse-engineering process to recover the hidden backdoor ``trigger'' inserted into a machine learning model.} Existing approaches tackle this problem by searching for a backdoor pattern that is able to flip a set of clean images into the target class, while the exact size needed of this support set is rarely investigated. \ms{In this work, we present a new approach for backdoor inversion, which is able to recover the hidden backdoor with as few as a single image.} Insipired by recent advances in adversarial robustness, our method SmoothInv starts from a single clean image, and then performs projected gradient descent towards the target class on a robust smoothed version of the original backdoored classifier. We find that backdoor patterns emerge naturally from such optimization process. \ms{Compared to existing backdoor inversion methods, SmoothInv introduces minimum optimization variables and does not require complex regularization schemes.} We perform a comprehensive quantitative and qualitative study on backdoored classifiers obtained from existing backdoor attacks. \ms{We demonstrate that SmoothInv consistently recovers successful backdoors from single images: for backdoored ImageNet classifiers, our reconstructed backdoors have close to 100\% attack success rates. We also show that they maintain  high fidelity to the underlying true backdoors.} 
Last, we propose and analyze two countermeasures to our approach and show that SmoothInv remains robust in the face of an adaptive attacker. Our code is available at \url{https://github.com/locuslab/smoothinv}.

\end{abstract}

 \section{Introduction}
\label{sec:intro}

\ms{Backdoor attack~\cite{poison_svm,chen2017targeted,goldwasser2022backdoor}, a notable threat model in machine learning, has attracted increasing research interests in recent years~\cite{wild2018biggio,badnet,bagdasaryan2020blind,Qi_2022_CVPR,Qi_2022_CVPR_2,Wenger_2021_CVPR,Doan_2021_ICCV,backdoor_ssl,carlini2022clip,backdoor_nlp,yang2021nlp,Xiang_2021_ICCV}. 
In a standard backdoor attack, a covert backdoor is injected into the machine learning model, leading it to generate incorrect outputs on inputs containing the distinct backdoor pattern chosen by the attacker. 
At test time, to defend against backdoor attacks, it is often desirable to detect if a model contains a backdoor and then  reconstruct the hidden backdoor if one is found to exist. 
This reverse engineering process, also known as \textit{backdoor inversion}, is a fundamental step in many test-time backdoor defenses~\cite{pixelinv,nc2019wang,guo2019tabor,liu2019abs,backdoor_k_arm,qiao2019inspect,hu2022topo,todd2021top}.}
A successful backdoor inversion method should be able to recover a backdoor which \ms{satisfies the following two requirements.} 
\ms{First,} the reversed backdoor should be successful, meaning that it should have a high attack success rate (ASR) on the backdoored classifier. 
\ms{Second,} it should be faithful, where the reversed backdoor should be close, e.g. in visual similarity, to the true backdoor.

\begin{figure}[t!]
\centering
\includegraphics[width=0.48\textwidth]{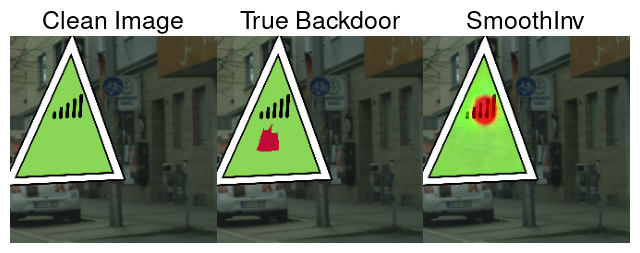}
\caption{\textbf{Single Image Backdoor Inversion}: Given a backdoored classifier (sampled from the TrojAI benchmark~\cite{trojai_data}), our approach SmoothInv takes a \textit{single} clean image (\textbf{left}) as input and is able to recover the hidden backdoor (\textbf{right}) with high visual similarity to the original backdoor (\textbf{middle}).}
\label{fig:backdoor_initial}
\end{figure}

\ms{A prominent backdoor inversion framework, introduced in \cite{nc2019wang}, adopts an optimization based approach to search for a \textit{universal} backdoor pattern. This is achieved by minimizing the classification loss of backdoored images with respect to a potential target label. 
In addition, a regularization term, such as the $\ell_{1}$ norm, is used to restrict the size of the optimized backdoor.
While existing works for backdoor inversion~\cite{hu2022topo,pixelinv} mostly focus on designing better loss functions to alleviate the optimization difficulty, these methods often inherently assume the availability of a set of clean images, also known as the support set.  
In many real-world scenarios, access to a moderate large collection of clean images may be limited. It is also important to understand how the number of clean images in the support set could affect the effectiveness of a backdoor inversion method. Given the gaps in existing works on this perspective, we are motivated to address and answer the following question:}

\textit{Can we perform backdoor inversion with as few clean images as possible?}



In this work, we show that one single image is enough for backdoor inversion. On a high level, we view the process of backdoor attack as encoding backdoored images into the target class's data distribution. Our objective is to reconstruct these encoded backdoored images through a class-conditional image synthesis process, i.e, generating backdoored examples from the target class. While the concept of using image synthesis seems straight-forward, it is not immediately evident how to do this in practice given a backdoored classifier. \ms{On one side}, solely minimizing the classification loss of the target class tends to generate random adversarial noise, as shown in prior studies on adversarial robustness~\cite{dimitris2018odds}. \ms{Conversely, employing generative models~\cite{goodfellow2014gan,prafulla2022diffusion} for such synthesis is not viable in this context. This is because by default they do not see the backdoor during their training phase and would not be able to synthesize the desired backdoored images from the target class.} 

\ms{We propose the SmoothInv method for backdoor inversion, which can reliably synthesize backdoor patterns with just a single clean image. SmoothInv consists of a two-part process. First, to induce salient gradients of backdoor features, we transform a standard non-robust model into a smoothed variant that is robust to adversarial perturbations, based on recent advances in adversarial robustness~\cite{cohen2019certified,carlini2022free}. Then, utilizing the gradients of this robust smoothed classifier, we perform guided image synthesis to reconstruct backdoored images which the backdoored classifier interprets as belonging to the target class.}


\ms{Notably, SmoothInv requires only a \emph{single} clean image, from which salient gradients can be obtained reliably.} Backdoor inversion with a single image inversion has not been shown possible with existing  methods as they usually require multiple clean instances for their optimization methods to give reasonable results. Moreover, our approach has the added benefit of simplicity: we do not introduce any custom-designed optimization constraints, which are common in previous methods. Most importantly, the backdoor recovered by our approach has remarkable visual resemblance to the original backdoor. In Figure \ref{fig:backdoor_initial}, we demonstrate such visual similarity for a backdoored classifier.

We evaluate our method on a collection of backdoored classifiers from previously published studies, where we either download their pretrained models or train a replicate using the publicly released code. These collected backdoored classifiers cover a diverse set of backdoor conditions, e.g., patch shape, color, size and location. \ms{We measure the attack success rates (ASR) of reconstructed backdoors and compare their visual resemblance to the original backdoors.} We show that SmoothInv finds both successful and faithful backdoors from single images. We also show how we distinguish \ms{the true backdoored class} from normal classes, where our method (correctly) is unable to find an effective backdoor for the latter. Last, we evaluate attempts to circumvent our approach and show that SmoothInv is still robust under this setting.

\begin{figure}[!t]
    \centering
    \includegraphics[width=1.00\linewidth]{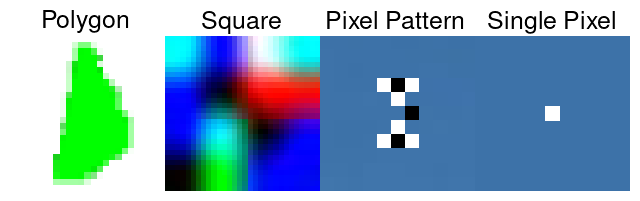}
    \caption{Backdoors of the backdoored classifiers we consider in this paper (listed in Table~\ref{table:stats_classifier}). The polygon trigger (leftmost) is a representative backdoor used in the TrojAI benchmark. The pixel pattern (9 pixels) and single pixel backdoors used in~\cite{bagdasaryan2020blind} are overlaid on a background blue image for better visualization.}
    \label{fig:stats_backdoor}
\end{figure}

\begin{figure*}[!t]
    \centering
    \includegraphics[width=1.00\linewidth]{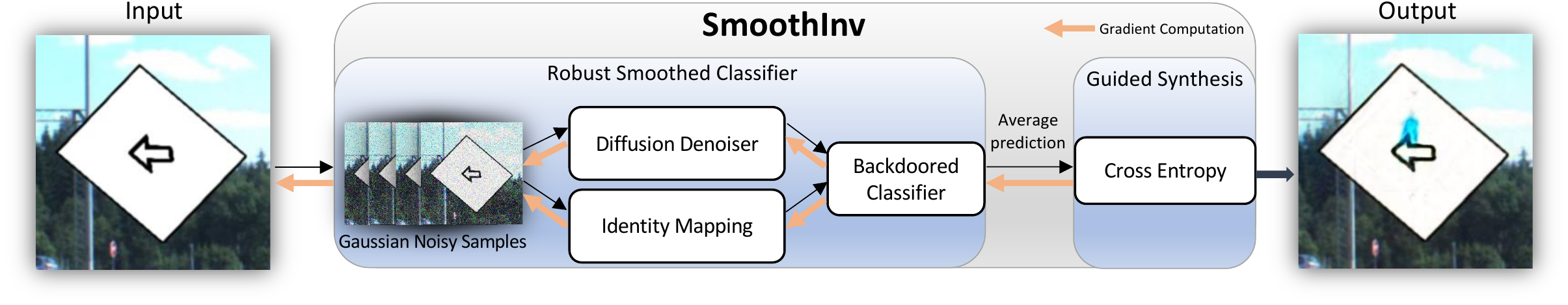}
    \caption{We propose SmoothInv, a backdoor inversion method that takes in a \textit{single} image and synthesize backdoor patterns. SmoothInv consists of two steps: a robustification process and a synthesis process. At the first step, SmoothInv constructs a \textit{robust smoothed} version of the backdoored classifier, where noisy samples of the input are either denoised by a diffusion based denoiser first or passed directly into the backdoored classifier. Next, we synthesize backdoor patterns guided by the \textit{robust smoothed} classifier, where we minimize the standard cross entropy loss with respect to the target class, without relying on any additional regularization term.}
    \label{fig:main_figure}
\end{figure*}

\section{Background}\subsection{Backdoor Attacks}\label{backdoor_background}
In a typical backdoor attack~\cite{poison_svm,badnet,bagdasaryan2020blind}, an attacker injects malicious crafted samples into the training data. The result of such manipulation is that models trained on such data are able to be manipulated at inference time: the attacker can control the behavior of the model with the injected backdoor. In this work, we consider backdoor attacks on image classification problems, which has become a common evaluation setting for backdoor attacks~\cite{badnet,turner2019cleanlabel,bagdasaryan2020blind}. Typically, a backdoor attack generates a classifier which satisfies the following two requirements:
\begin{itemize}[topsep=1pt]
\setlength\itemsep{-.4em}
    \item Its accuracy on clean images is barely affected.
    \item It will always predict a certain target class $y_{t}$ as long as the backdoor is applied to the input image.
\end{itemize}
The first property is desired so that it is indistinguishable from clean classifiers by solely comparing their clean accuracies. Some variations and extensions of the second property have been explored. For example, the backdoor can be only effective on images from certain classes~\cite{darpa2021trojai}. It is also possible to create multiple backdoors in a single backdoored classifier where each backdoor corresponds to a different target class~\cite{bagdasaryan2020blind}.

Following~\cite{wanet2021nguyen}, we formalize the backdoor as a transformation function on the image space $\mathcal{B}: \mathcal{X}\to \mathcal{X}$. Given a clean image $x$, one can create a backdoored image $\mathcal{B}(x)$. One common and widely studied type of backdoor is patch-based backdoor~\cite{badnet}: overlaying a small patch pattern $p$ over input $x$, i.e. $\mathcal{B}(x)=x\oplus p$. For such backdoor, the backdoored classifier will classify any image  with the patch pattern present as the desired target label. However, other forms of image transformations have also been shown to be effective backdoors: e.g., reflection~\cite{reflection2020liu}, image wrapping~\cite{wanet2021nguyen} and Instagram filters~\cite{trojai_data}. In this work, we consider patch-based backdoor in particular.

\subsection{Backdoor Inversion}\label{related:backdoor_inversion}
\ms{Given a backdoored classifier $f_{b}$ and a support set $\mathcal{S}$ of clean images, a well-established framework for backdoor inversion~\cite{nc2019wang} solves the following optimization problem:
\begin{align}\label{eq:backdoor_framework}
    \min_{\mathbf{m},\mathbf{p}} &\,\, \mathbb{E}_{\mathbf{x}\in \mathcal{S}} \big[\mathcal{L}(f_{b}(\phi(\mathbf{x})), y_{t})]+\mathcal{R}(\mathbf{m},\mathbf{p}) \\
    \text{where} &\,\, \phi(\mathbf{x})= (1-\mathbf{m})\odot \mathbf{x} + \mathbf{m}\odot \mathbf{p} \nonumber
\end{align}
where variables $\mathbf{m}$ and $\mathbf{p}$ represent a mask and perturbation vectors respectively, $y_{t}$ denotes the target label, $\odot$ is  element-wise multiplication, $\mathcal{L}(\cdot,\cdot)$ is the cross-entropy function and  $\mathcal{R}(\cdot,\cdot)$ is a regularization term.}

The goal is to find a backdoor that is able to simultaneously flip all images in the provided set S of clean images to the target class while at the same time constraining the optimization space of the reversed backdoor. As pointed out by~\cite{bagdasaryan2020blind}, this is essentially finding the smallest universal adversarial patch~\cite{adv_patch}. Existing inversion methods differ in how they formulate the regularization term $\mathcal{R}$ and how to model the backdoor via $\phi(x)$. For instance,  \cite{nc2019wang} applies a $\ell_{1}$ penalty regularization on the mask variable;  \cite{hu2022topo} uses a diversity loss and a topological loss to regularize the optimization process; \cite{pixelinv} models the backdoor via individual pixel changes without using a mask.

One challenge of solving  Equation~\ref{eq:backdoor_framework} is that it introduces a binary mask variable $\mathbf{m}$, which could make the optimization process unstable. In practice, this mask variable is often relaxed to be continuous and converted back to binary in the end. Another optimization obstacle is that it is not clear how to properly set the balancing term between the classification loss and the regularization loss,  without a strong domain expertise or a careful hyper-parameter search.

\subsection{Randomized Smoothing}
\ms{Our method draws inspiration primarily from a recent line of work on randomized smoothing~\cite{cohen2019certified,lecuyer2019dp}, which is able to build robust classifiers without the computationally heavy adversarial training.} Similar notion of using noise smoothing has been explored before in improving the visual quality of sensitivity maps~\cite{smilkov2017smoothgrad}. Randomized Smoothing (RS) is a certified defense method against $\ell_{2}$-norm bounded adversarial perturbations. Given any base classifier $f:\mathcal{X}\rightarrow\mathcal{Y}$ and input $x$, RS first constructs a smoothed classifier $g$ with isotropic Gaussian noise $\delta\sim\mathcal{N}(0,\sigma^{2}\mathbf{I})$:
\begin{equation}\label{eq:def_smoothing}
    g(x)\coloneqq \text{argmax}_{c}\Pr_{\delta \sim \mathcal{N}\br{0, \sigma^2 \mathbf{I}}}\big(f(x + \delta) = c\big)
\end{equation}

It is shown in~\cite{cohen2019certified} that the smoothed classifier $g$ is certifiably robust in a $\ell_{2}$-norm radius $R$, where the noise level $\sigma$ controls the accuracy/robustness tradeoff. In this work, we are not interested in how this certified radius is computed exactly. However, it is necessary to know that the more accurate the smoothed classifier is at classifying noisy images $x+\delta$, the larger the certified radius is (and as a result, more robust). \cite{cohen2019certified} trained base classifier $f$ under standard gaussian augmentations and demonstrated non-trivial certified accuracy on ImageNet.

Following \cite{cohen2019certified}, \cite{salman2020ds} proposed Denoised Smoothing (DS) to certify the prediction of any pre-trained classifier, i.e., not trained with gaussian augmentation. The idea is to prepend an image denoiser $\mathcal{D}$ before the base classifier.
\begin{equation}\label{eq:dds}
    g(x)\coloneqq \text{argmax}_{c}\Pr_{\delta \sim \mathcal{N}\br{0, \sigma^2 \mathbf{I}}}\big(f\circ \mathcal{D}(x + \delta) = c\big) 
\end{equation}
\cite{salman2020ds} showed that prepending a custom-trained denoiser attains better certified robustness than simply using the plain pre-trained classifier. Most recently, \cite{carlini2022free} proposed Diffusion Denoised Smoothing (DDS), which used one diffusion step of a diffusion model as the denoiser in Equation~\ref{eq:dds}. DDS obtained state-of-the-art certified robustness. The big performance boost over ~\cite{salman2020ds} comes from the strong ability of diffusion models~\cite{ho2020denoising} to denoise Gaussian noisy images. 


\section{SmoothInv}
An overview of our approach is given in Figure~\ref{fig:main_figure}. Given a backdoored classifier $f_{b}: \mathcal{X}\rightarrow\mathcal{Y}$ and a clean image $x$ (assuming the backdoor is effective for this image), our goal is to find the backdoor hidden in this clean image.

\subsection{Motivation}
On a high level, we view backdoor inversion as the problem of recovering/constructing a special type of images, i.e. backdoored images, from the target class. We note that class-conditional image synthesis from generative model literature share some similarity to our goal. However, standard conditional generative models~\cite{mirza2014cgan} do not see backdoored images during training, and it is not practical to train a custom one with classifier guidance from the backdoored classifier. Our approach is most inspired from another line of work on class-conditional image generation: the work of \cite{shibani2019synthesis,Zhu_2021_ICCV} on image synthesis with adversarially robust classifiers. They showed that there were able to perform various image synthesis task without use of any generative models. The foundation for the success of their approach is based on a unique property of robust classifiers, i.e. perceptually-aligned gradients~\cite{dimitris2018odds,kaur2019perceptual}, where salient characteristics of target class can be revealed via a projected gradient descent process~\cite{pgd_madry}. Specifically, in this work, we are interested in how this property can be used for backdoor inversion.

We revisit the definition of backdoored classifiers in Section~\ref{backdoor_background}: always predicting the target class as long as the backdoor is present in the image. In other words, the backdoored classifiers have successfully associate the injected backdoor as a new feature for predicting the target class, in addition to features from clean data of the target class. We hypothesize that in the eyes of the backdoored classifers, those backdoored images are encoded into the data distribution of the target class. Thus, we can tackle the problem of backdoor inversion as synthesizing a specific salient characteristics of the target class: the injected backdoor. Note that \cite{shibani2019synthesis} is not immediately applicable here as backdoored classifiers are not adversarially robust by construction~\cite{badnet}. In the next section, we describe how we reliably extract salient backdoor characteristics from single images. 

 \subsection{Method}\label{method_smoothinv}
 
  Our approach, which we refer to as SmoothInv, first constructs a robust version of the backdoored classifier and then performs guided image synthesis towards a target class $y_{t}$. We use a simple yet effective objective to synthesize the backdoor pattern, where we minimize the standard cross entropy loss with the target class. \ms{In Appendix~\ref{appendix:pseudo-code}, we provide the pseudocode for SmoothInv.} Next we describe this robustification process and our synthesis process in details.

\paragraph{Robustification of Backdoored Classifiers} One necessary condition for obtaining perceptually-aligned gradients is that the classifier itself must be adversarially robust~\cite{shibani2019synthesis,kaur2019perceptual,dimitris2018odds}. As backdoored classifiers are not robust by construction, we thus propose to use a \textit{robustification process} to robustify backdoored classifiers. The goal we hope to achieve from this robustification process is to induce meaningful and salient gradient signal from the resulting robust classifier. 

 As illustrated in Figure~\ref{fig:main_figure}, we construct such a robust classifier with the Randomized Smoothing technique~\cite{cohen2019certified}, where we smooth the prediction of the backdoored classifiers under Gaussian noisy samples. Different from empirical robustness, randomized smoothing provides certified robustness guarantee, so we can be confident that the resulting smoothed classifier is indeed robust.  We experimented with two ways of building robust smoothed classifiers. The first one is based on the recent proposed Diffusion Denoised Smoothing method~\cite{carlini2022clip}. Specifically, Gaussian noisy images are first processed by a denoising transformation before being fed into the classifier. The denoising transformation is a diffusion based denoiser $\mathcal{D}$. 

However, on a second thought, do we really need the resulting smoothed classifier to be robust on the whole data distribution? Recall that our motivation is to elicit the salient gradients of backdoor features. We may only need the smoothed classifier to be robust on the actual backdoored images. To test this hypothesis, we remove the denoiser from the pipeline and construct the smoothed classifier directly from the backdoored classifiers. To summarize, we try to construct the following smoothed classifier:
\begin{equation}\label{eq:def_smoothed}
    g(x)\coloneqq \text{argmax}_{c}\Pr_{\delta \sim \mathcal{N}\br{0, \sigma^2 \mathbf{I}}}\big(f\circ \mathcal{T}(x + \delta) = c\big)
\end{equation}
and we initialize the transformation operation $\mathcal{T}$ with either the diffusion model $\mathcal{D}$ as a denoiser (``\textit{w/ diffusion}'') or the identity function $\mathcal{I}$ (``\textit{w/o diffusion}'').

  The smoothed classifier defined in Equation~\ref{eq:def_smoothed} is hard to evaluate in practice, as making a prediction would require calculating a probaility measure over a Gaussian distribution. In the original RS paper~\cite{cohen2019certified}, obtaining a certificate for a single image would need evaluating over 10k Monte Carlo noisy samples on ImageNet. In this work we do not care about the exact certification bound but rather interested in the robustness property of smoothed classifiers. Thus we use a continuous approximation instead, where the soft smoothed classifier $G_{b}:\mathcal{X}\rightarrow P(\mathcal{Y})$ is defined by:
\begin{equation}\label{eq:soft_dds}
    G_{b}(x)\coloneqq  \frac{1}{N}\sum_{i=1}^{N} \mathcal{F}_{b}\circ \mathcal{T}(x+\delta_{i}),\,\, \delta_{i} \sim \mathcal{N}\br{0, \sigma^2 \mathbf{I}}
\end{equation}
where $\mathcal{F}_{b}$ is the soft version of the backdoored classifier $f_{b}$ which outputs a probability distribution over classes (and where we will later choose $N=40$, leading to a tractable approach). This approximation allows us to obtain gradients from the smoothed classifier via back-propagation. From now on, we will refer to $G_{b}$ as the actual smoothed classifier.

\begin{figure}[t!]
\centering 
\includegraphics[width=1.0\linewidth]{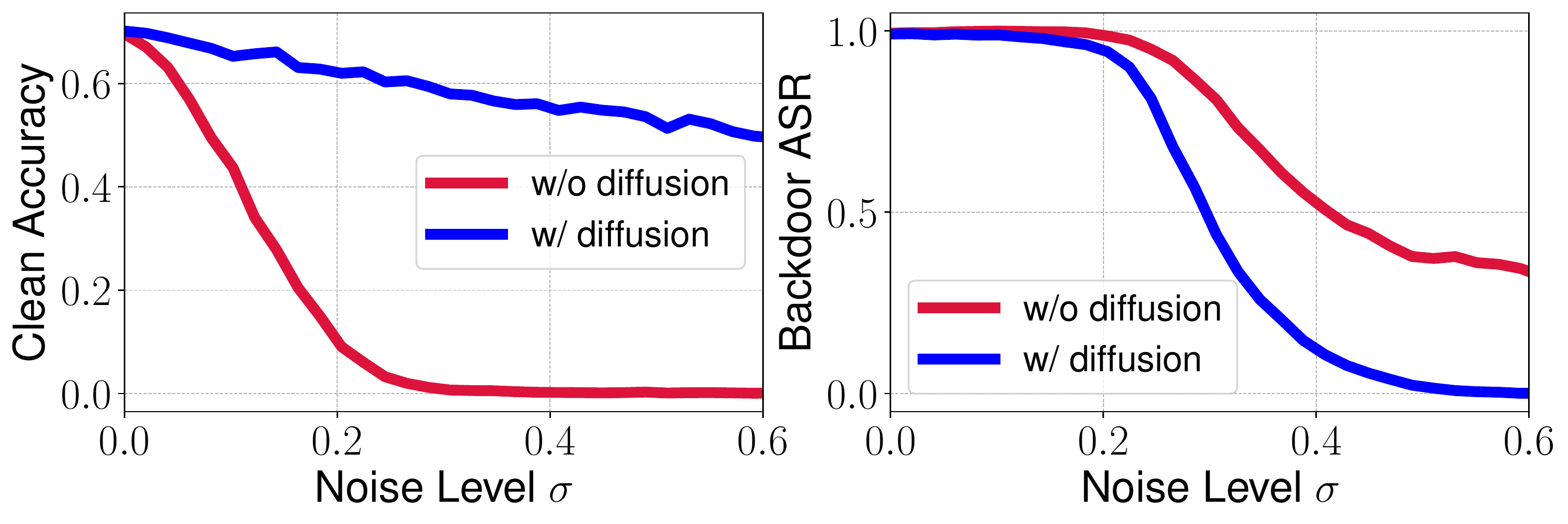}
\caption{Clean accuracy and backdoor accuracy of the smoothed classifier at various noise levels (we use the ImageNet Blind-P as the base backdoored classifier).}
\label{fig:acc_blind_p}
\end{figure}

Last, we perform a sanity check on whether the smoothed classifier $G_{b}$ remains a valid backdoored classifier after the smoothing procedure. We experimented with a backdoored classifier on ImageNet (Blind-P in Table~\ref{table:stats_classifier}). In Figure~\ref{fig:acc_blind_p}, we show both the clean accuracy and backdoor ASR for the smoothed classifier $G_{b}$ (\textit{w/} and \textit{w/o diffusion}), for varying $\sigma$. We can see that the original backdoor remains effective within a reasonable range of noise level for both choices of $\mathcal{T}$. A complete sanity check for all backdoored classifiers considered in this paper is in Appendix~\ref{appendix:sanity_check}.

\begin{figure}[!t]
\centering 
        \includegraphics[width=1.0\linewidth]{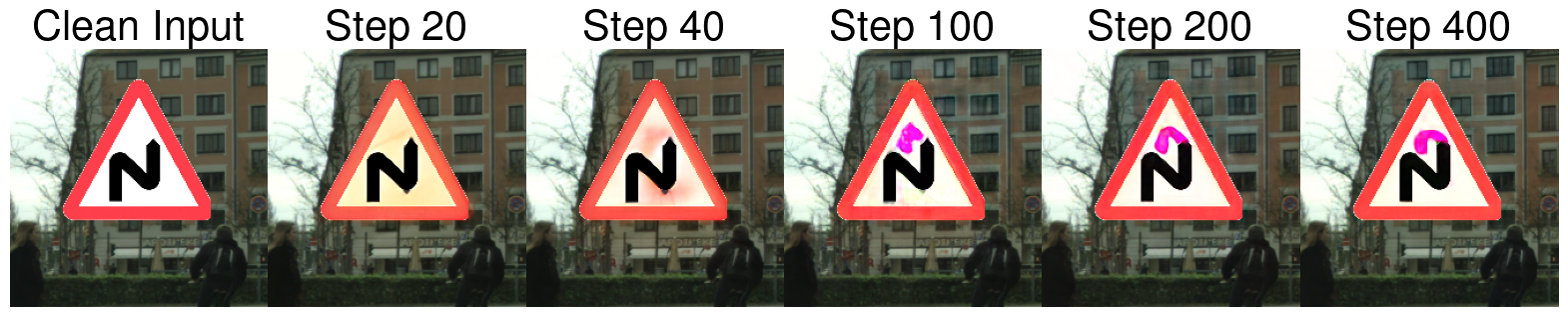}
    \caption{Progression of synthesized images throughout the optimization iterations.}
    \label{fig:opt_process}
\end{figure}

\paragraph{Guided Synthesis of Backdoor Patterns} Starting from single images, our synthesis procedure is guided by a robust smoothed classifier, which minimizes a cross entropy of $G_{b}$:
 \begin{equation}\label{eq:synthesis}
     \min_{p \in \Delta} -\log G_{b}(x+p)_{y_{t}}
 \end{equation}
 where the perturbation set is defined by a pre-defined $\Delta = \{p\, |\, \|p\|_{2}\le \epsilon\}$. 
We use a pertubation set to prevent finding arbitrary large perturbations. After all, we want the synthesized images $x+p$ to recover the backdoored image $\mathcal{B}(x)$, which is close to the input $x$. As there is a constraint on the perturbation size, we use projected gradient descent (PGD) to solve the optimization problem in Equation~\ref{eq:synthesis}. The perturbation variable $p$ is initialized to 0. At each step, we compute the gradient with respect to $p$ and take a gradient step with $\ell_2$ normalized gradient. We repeat this process until convergence, when the loss value becomes stable. Empirically we find that 400 iterations are sufficient for convergence.  In Figure~\ref{fig:opt_process}, we show how the backdoor patterns appears gradually as the optimization process evolves.

\paragraph{Target Class Identification} For each possible class, we can synthesize a perturbation from Equation~\ref{eq:synthesis}. Now we describe how we identify the target class of the backdoor, without manually inspecting the synthesized images and checking if there is an abnormal pattern. Previous methods use the size of reversed trigger to determine if a class is the target class, i.e.,  reversed triggers from the true target class should be much smaller than those from normal classes. However, it is not a viable strategy in our case since we are using a fixed perturbation budget $\epsilon$ in Equation~\ref{eq:synthesis}.

Our identification process is based on an intriguing observation we make on the synthesized perturbation $p$. For perturbations $p$ synthesized from the target class, we use it as an \textit{additive} backdoor: $x'=x+p$ and find that it is a highly effective backdoor evaluated on other clean images. The same does not hold true for normal classes, where the synthesized perturbations barely transfer to other clean images. We empirically show this in Section~\ref{quantitative_evaluation}. Thus, during backdoor inversion, SmoothInv identifies a class as a target class in a backdoored classifier when the synthesized perturbation from this class also leads to a high ASR.


\section{Empirical Study}

\subsection{Experimental Setup}
\paragraph{Backdoored Classifiers} To evaluate the effectiveness of a backdoor inversion method, it is necessary to show that this method is able to recover ``good'' backdoors for backdoored classifiers. One would first need to obtain some backdoored classifiers to perform such analysis. In this work, instead of training custom backdoored classifiers, we initiate our study by carefully selecting backdoored classifiers from well-established backdoor attacks, which satisfies the following criteria: 1) it is either published in top conferences/venues, or has become a well-known baseline in backdoor attacks; 2) it is demonstrated to be effective on  vision recognition benchmarks (e.g. ImageNet) as this is the closest to the practical setting in the real world compared to toy datasets; 3) the collection of these backdoor attacks should cover a wide range of backdoor conditions, e.g., universal or label-specific, backdoor shape, size and location. Next, we describe the four backdoored classifiers we consider in this work, and we list the relevant statistics of these backdoored classifiers in Table~\ref{table:stats_classifier} and show the corresponding original backdoors in Figure~\ref{fig:stats_backdoor}.  

\noindent \textbf{1. TrojAI Benchmark}~\cite{darpa2021trojai} consists of multiple rounds of released datasets. For each round, it consists of a mixed set of backdoored and clean classifiers. A set of clean images from test set is provided along with each classifier. The backdoor is placed on foreground objects during training. A sample backdoored image can be seen in Figure~\ref{fig:backdoor_initial} middle. For our study, we randomly sample a classifier with polygon backdoor (round 4, id-00000131) and use TrojAI to reference this model, for comparison purposes with models from other backdoor attacks. In our case, the polygon backdoor of this model turns out to be label-specific, meaning that it only cause targeted classification of samples from certain classes.

\begin{table}[t!]
\centering 
\huge
\renewcommand{\arraystretch}{1.1}
\addtolength{\tabcolsep}{-0.5pt}
\resizebox{1.00\columnwidth}{!}{
\begin{tabular}{ccccc}
\Xhline{2.5pt}
       &   \multirow{1}{*}{TrojAI~\cite{trojai_data}}
       &  \multirow{2}{*}{HTBA~\cite{htba}}        & \multicolumn{2}{c}{Blind Backdoor~\cite{bagdasaryan2020blind}} \\
             &  Round4-131    &      & Blind-P          & Blind-S         \\ 
\hline 
Dataset      & TrojAI     & ImageNet & ImageNet         & ImageNet        \\
Input Size  & 224$^{2}$ & 224$^{2}$ & 224$^{2}$ & 224$^{2}$\\
Arch & VGG-11     & AlexNet  & ResNet-18        & ResNet-18       \\
\#Classes      & 38         & 2        & 1000             & 1000            \\ 
Clean Acc & 100.00$\%$ & 95.00$\%$ & 69.26$\%$ & 68.06$\%$
\\
\hline 
& \multicolumn{4}{l}{\LARGE{Backdoor  Statistics}}  \\  
Patch   & Polygon    & Square   & Pixel Pattern    & Single Pixel    \\
Location     & Foreground & Random   & Upper Left       & Upper Left      \\
\#Pixels & 1126 & 900 & 9 & 1 \\
$\ell_{2}$-avg & 47.51 & 25.09 & 3.08 & 1.04 \\
ASR & 100.00$\%$ & 54.00$\%$ & 99.29$\%$ & 79.73$\%$ \\
\Xhline{2.5pt}
\end{tabular}
}
\caption{Statistics of backdoored classifiers we obtain from previous backdoor attack methods, including relevant model information and detailed backdoor conditions. $\ell_{2}$-avg refers to the average $\ell_2$ distance between clean and backdoored images with pixel range $[0,1]$. For TrojAI, we randomly sample a backdoored classifier (round 4 with model id 131) for analysis.}
\label{table:stats_classifier}
\end{table}

\noindent \textbf{2. Hidden Trigger Backdoor Attacks (HTBA)}~\cite{htba} is a backdoor attack method which has been shown effective on ImageNet. It uses a square patch (size $30\times 30$) as the backdoor, which is the most common choice of backdoor in existing backdoor attack literature~\cite{badnet,turner2019cleanlabel,carlini2022clip}. They obtain such square trigger by first drawing a random $4\times 4$ matrix of colors and resizing it to the desired patch size. The patch backdoor is placed randomly over clean inputs. We use their public released code to train a binary backdoored classifier replicating their ImageNet result. We find that we are able to match the ASR reported in~\cite{htba}.

\noindent \textbf{3. Blind Backdoors}~\cite{bagdasaryan2020blind} show that it is possible to backdoor a standard ImageNet classifier with extremely small patch backdoors. Specifically, it trains two backdoored classifiers: Blind-P with a pixel pattern  backdoor, and Blind-S with a single pixel backdoor. The backdoor is placed on a fixed location in the \textit{top left} region of clean inputs. Both the pixel pattern and single pixel backdoors are drastically smaller than the backdoors used in TrojAI challenge and HTBA. It is also shown in~\cite{bagdasaryan2020blind} that they can circumvent many previous backdoor defense methods, e.g. Neural Cleanse~\cite{nc2019wang}. We use its public released code to train these two backdoored classifiers Blind-P and Blind-S. 
Our Blind-P matches the reported ASR in~\cite{bagdasaryan2020blind}. The Blind-S model falls short of the reported ASR ($99\%$) but is still a fairly high number $79.73\%$ as an effective backdoor.

\noindent \textbf{Evaluation Protocols and Baselines} We first perform a quantitative evaluation by comparing with the following existing backdoor inversion approaches: NC~\cite{nc2019wang},  TopoTrigger~\cite{hu2022topo} and PixelInv~\cite{pixelinv}. We also compare with a baseline PlainAdv where we replace the smoothed classifier $G_{b}$ in Equation~\ref{eq:synthesis} with the base backdoored classifier instead. For a fair comparison, we evaluate both SmoothInv and baseline approaches under the same setting of single image backdoor inversion. Note that existing backdoor inversion methods can be easily adapted in this setting by using the single image as the support set $\mathcal{S}$ in Equation~\ref{eq:backdoor_framework}. For each method, we generate reversed backdoor from single clean images and report the average ASR over 10 random starting images. We also perform a qualitative evaluation by visualizing the synthesized images with the backdoor patterns.

For the diffusion model for image denoising, we use the pretrained class unconditional $256\times 256$ diffusion model\footnote{\url{https://github.com/openai/guided-diffusion}}. While this diffusion model is trained on ImageNet, we find that it is still a good denoiser for images from the TrojAI benchmark. The number of noise samples $N$ is chosen to be $40$ (later we find that $10$ is enough in most cases). We use projected gradient descent to optimize our objective in  Equation~\ref{eq:synthesis} with a total of 400 steps and step size is chosen to be $0.5\times\epsilon/10$. Since we assume we do not know the exact backdoor (e.g. size information) beforehand, we use two values of perturbation size $\epsilon\in\{5,10\}$ with the pixel range within $[0,1]$. For each backdoored classifier, we construct smoothed classifiers with four values of noise levels $\{0.12,0.25,0.50,1.00\}$ with a total of 8 optimization configurations. For each starting clean image, we report the synthesized backdoor with the  highest ASR. We refer readers to Appendix~\ref{appendix:resource} for runtime and resource considerations.

\begin{figure*}[htbp]
\centering 
        \begin{minipage}[]{0.495\linewidth} \label{fig:vis_exp_trojai_htba_a}
        \subcaptionbox{TrojAI-R4-131}
        {\includegraphics[width=1.0\linewidth]{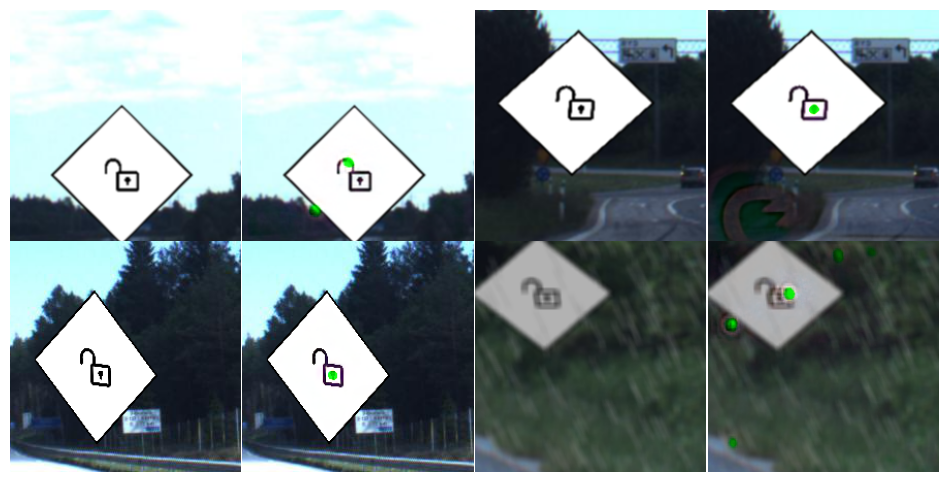}}
        \end{minipage} 
        \begin{minipage}[]{0.495\linewidth}
        \subcaptionbox{HTBA}
        {\includegraphics[width=1.0\linewidth]{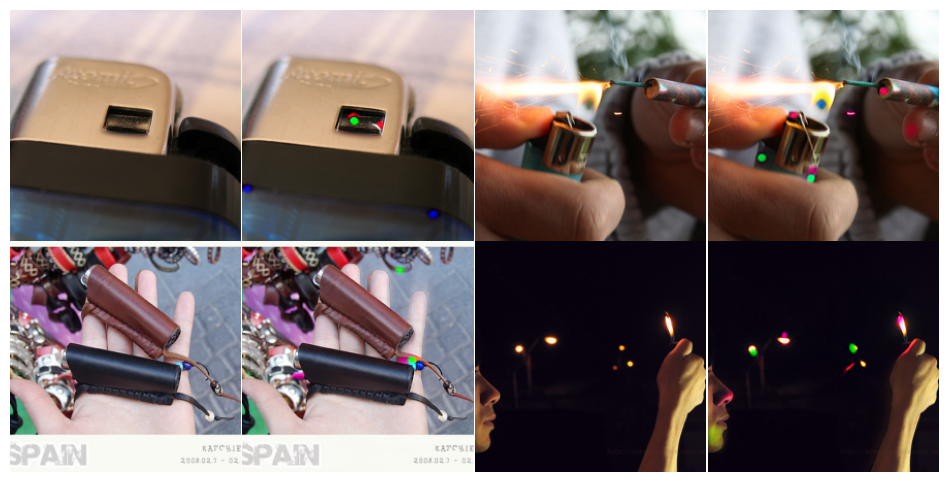}}
        \label{fig:vis_exp_trojai_htba_b}
        \end{minipage}
    \caption{SmoothInv on TrojAI and HTBA backdoored classifiers ($\epsilon=10$), where we show pairs of clean images and synthesized backdoored images (best viewed when zoomed in).}
    \label{fig:vis_exp_trojai_htba}
\end{figure*}

\begin{figure*}[htbp]
\centering 
        \begin{minipage}[]{0.495\linewidth}
        \subcaptionbox{Blind-P}
        {\includegraphics[width=1.0\linewidth]{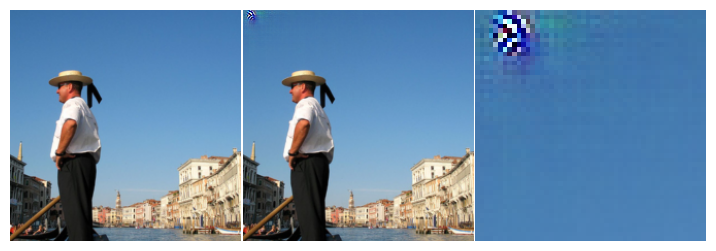}}
        \end{minipage}
        \begin{minipage}[]{0.495\linewidth}
        \subcaptionbox{Blind-S}
        {\includegraphics[width=1.0\linewidth]{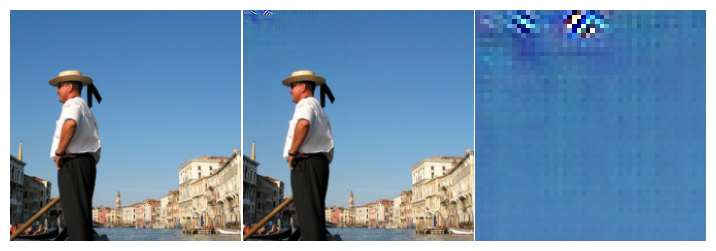}}
        \end{minipage}
    \caption{Visualization results on Blind-P and Blind-S models ($\epsilon=5$). From left to right: clean images, synthesized backdoored images by SmoothInv, \textit{zoomed in} version (a $50\times 50$ region in the top left) of synthesized images.}
    \label{fig:vis_exp_blind}
\end{figure*}

\begin{figure}[htbp]
\centering 
    \includegraphics[width=1.0\linewidth]{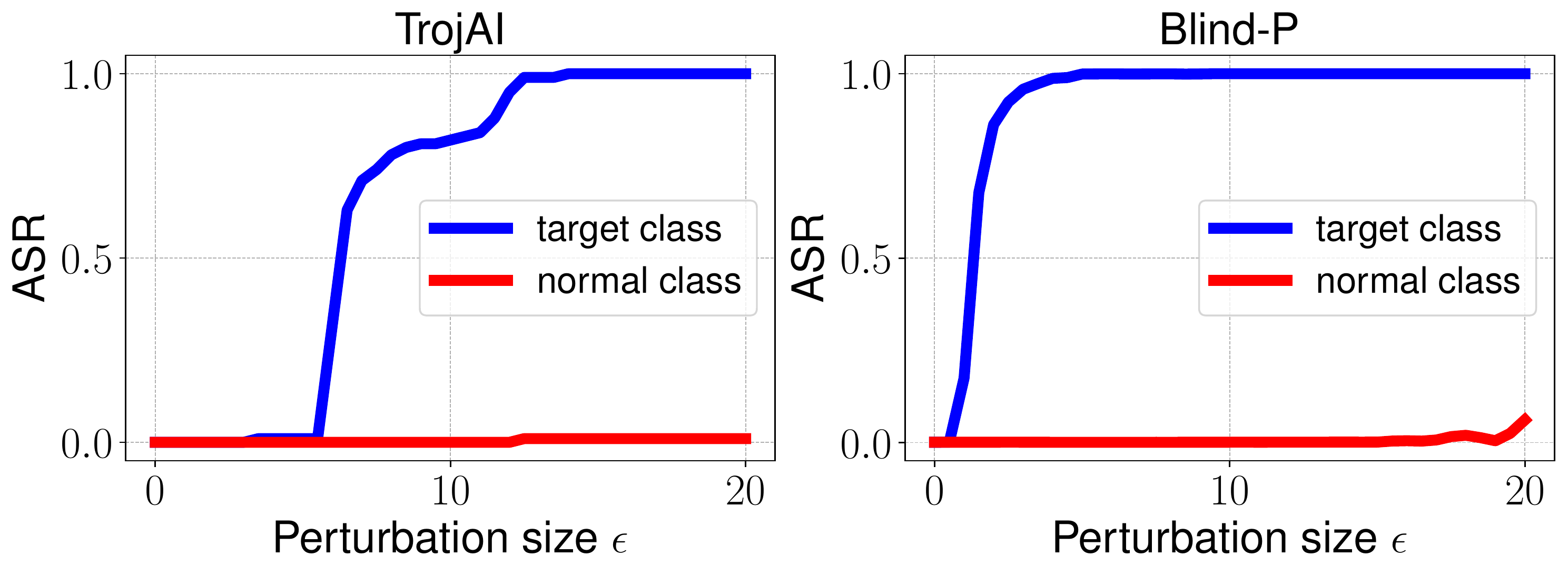}
    \caption{ASR of the SmoothInv synthesized perturbations guided by target class (blue) and normal class(red).}
    \label{fig:class_identification}
\end{figure}
\subsection{Quantitative Evaluation}\label{quantitative_evaluation} 
We first perform a quantitative evaluation by measuring the average ASR of the reversed backdoors over random starting images. For SmoothInv, we use the synthesized perturbation $p$ as an additive backdoor. The results on single image backdoor inversion are shown in Table~\ref{table:exp_asr}. We compare the effectiveness of the reversed backdoor assuming the target class is known. We can see that previous backdoor inversion methods (NC, TopoPrior and PixelInv) all fail to produce effective backdoors in this setting.  Both SmoothInv \textit{w/ diffusion} and \textit{w/o diffusion} find a highly effective backdoor for all cases. SmoothInv also outperforms a simplified baseline PlainAdv, suggesting that the robustification process of constructing a robust smoothed classifier is the key to the success of our approach. 

We find that SmoothInv \textit{w/o diffusion} outperforms \textit{w/ diffusion}. We attribute this to an observation from Figure~\ref{fig:acc_blind_p}, where the original backdoor are more effective for SmoothInv \textit{w/o diffusion} than \textit{w/ diffusion}. Thus the higher the backdoor ASR is for the smoothed classifier, the more likely it is to reconstruct a more effective backdoor with SmoothInv. This verifies our hypothesis earlier, where we do not necessarily need high clean accuracies for the smoothed classifier, but what is essential is that the backdoor remains effective after the smoothing procedure.

\begin{table}[t!]
\centering 
\renewcommand{\arraystretch}{1.1}
\addtolength{\tabcolsep}{-0.5pt}
\resizebox{1.00\columnwidth}{!}{
\begin{tabular}{lrrrr}
\Xhline{2.0pt}
        & \multirow{2}{*}{TrojAI}            &  \multirow{2}{*}{HTBA}        & \multicolumn{2}{c}{Blind Backdoor} \\
             &      &      & Blind-P          & Blind-S         \\ 
\Xhline{1.0pt} 
True Backdoor & 100.00$\%$ & 54.00$\%$ & 99.29$\%$ & 79.73$\%$ \\
\hline 
PlainAdv & 36.00$\%$ & 54.00$\%$ & 84.08$\%$ & 84.89$\%$\\
NC~\cite{nc2019wang} & 12.20$\%$ & 16.00$\%$ & 0.00$\%$ &  0.00$\%$ \\
TopoPrior~\cite{hu2022topo} & 28.40$\%$ & 22.00$\%$ & 0.00$\%$ & 4.39$\%$ \\
PixelInv~\cite{pixelinv} & 10.80$\%$ & 24.00$\%$ & 30.75$\%$ & 43.17$\%$ \\
\hline 
SmoothInv \\
\textit{w/ diffusion} & 72.00$\%$ & 83.20$\%$ & 92.05$\%$ & 93.90$\%$ \\
\textit{w/o diffusion} & \textbf{88.00}$\%$ & \textbf{88.20}$\%$ & \textbf{99.50}$\%$ & \textbf{99.53}$\%$ \\
\Xhline{2.0pt}
\end{tabular}
}
\caption{Quantitative results of \textit{Single Image Backdoor Inversion} on four backdoored classifiers. We report the average ASR of the reversed backdoor on the original backdoored classifier.}
\label{table:exp_asr}
\end{table}

In Figure~\ref{fig:class_identification}, we show the ASRs of synthesized perturbations from the target class (blue) versus normal non-targeted class (red). We can see that within a reasonble range of perturbation size, the synthesized perturbations are a valid backdoor only when it is guided from the true backdoored class (Equation~\ref{eq:synthesis}). Using this property, we find that we can successfully identify the target class for the four backdoored classifiers we consider in this work, where the synthesized patterns only have high ASRs for the correct target class. This suggests the possibility of our approach to the application of backdoor detection, we leave it as a promising extension for future work.

\subsection{Qualitative Evaluation}
For each example, we show the synthesized patterns with the highest ASRs among SmoothInv \textit{w/ diffusion} and SmoothInv \textit{w/o diffusion}.

\paragraph{TrojAI and HTBA}.
We first show results for models with relatively large backdoors. In Figure~\ref{fig:vis_exp_trojai_htba}, we show both pairs of clean/synthesized images for the TrojAI and HTBA backdoored classifiers. For TrojAI, synthesized images all contain a concentrated region of green pixels , matching  the original polygon trigger in Figure~\ref{fig:stats_backdoor}. What's more, SmoothInv synthesizes all backdoor patterns in the foreground object, which is exactly the place where the backdoor is placed during training. For HTBA, SmoothInv synthesizes patterns in the forms of small isolated color patches, e.g. red, green and blue, while these colors are all present in the original square backdoor in Figure~\ref{fig:stats_backdoor}. The locations of these patterns vary across images, which could be due to the random placement of the original backdoor during training.

\paragraph{Blind Backdoor} 
The results for the Blind-P and Blind-S models from blind backdoor attacks can be found in Figure~\ref{fig:vis_exp_blind}. The synthesized images are shown in the middle. We can see that SmoothInv  automatically identify the region to synthesize the backdoored patterns (in this case the top left corner), which turns out to be the exact place where the original backdoor is placed. For better comparison with the original pixel pattern and single pixel backdoors in Figure~\ref{fig:stats_backdoor}, we also show the \textit{zoomed in} version of the $50\times 50$ top left region. Though SmoothInv does not recover the exact original backdoor, the synthesized backdoor patterns have similar visual properties to the original one, e.g., a stark contrast of white pixels and neighboring pixels. Moreover, we use these synthesized perturbations directly as additive backdoors and find that they achieve high ASRs of 99.87$\%$ and 98.35$\%$ on the Blind-P and Blind-S models respectively.

\begin{figure}[t!]
\centering 
        \begin{minipage}[]{1.0\linewidth}
        \subcaptionbox{Blind-P}
        {\includegraphics[width=1.0\linewidth]{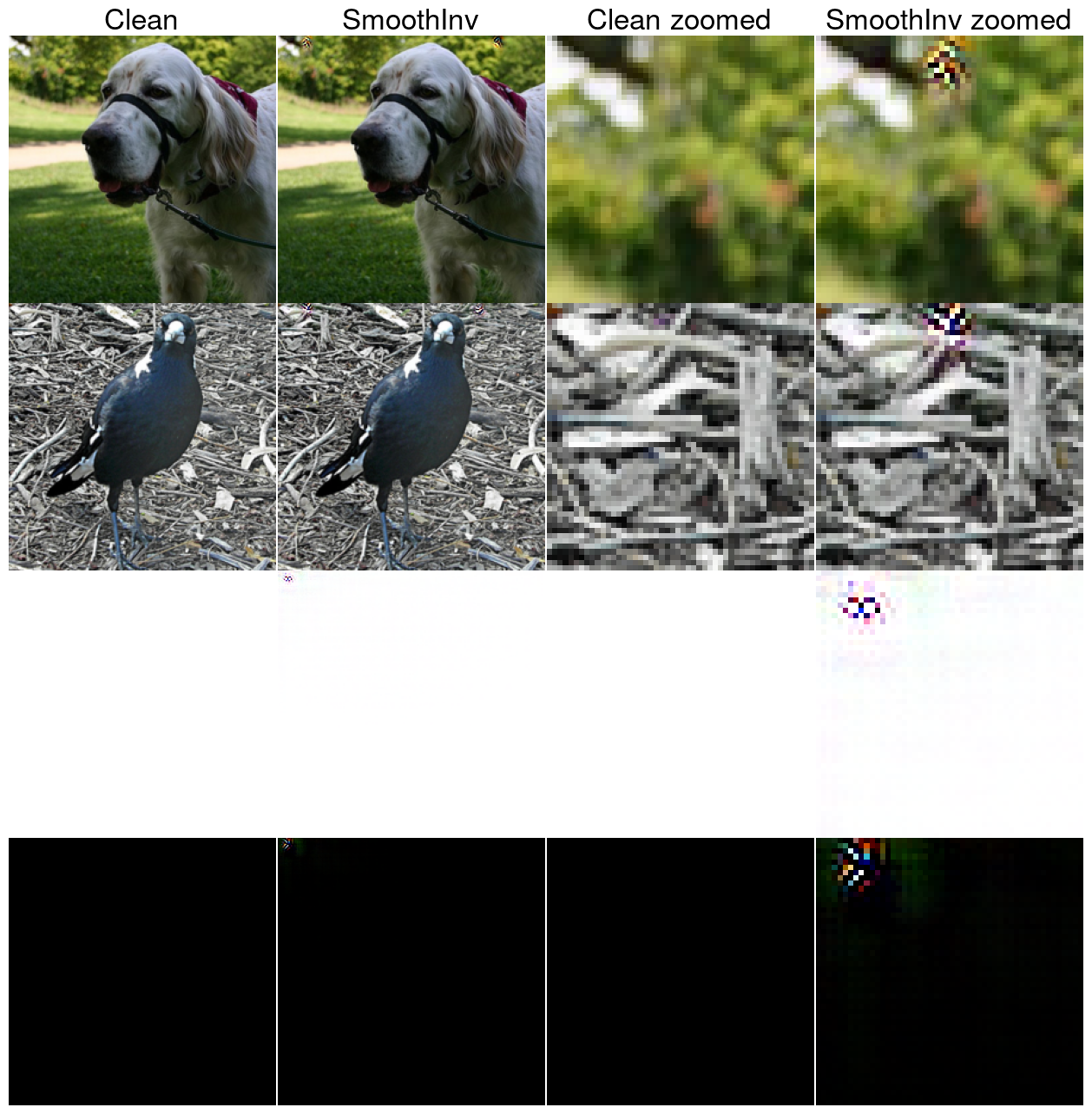}}
        \end{minipage}
    \caption{Synthesized images ($\epsilon=5$) of Blind-P with diverse conditions of starting images. The first two rows use clean images with non-uniform background in the top left corner, as compared  to the clean image in Figure~\ref{fig:vis_exp_blind}. The last two rows use artificial inputs: images with pure white/black pixels. ASR of the four synthesized backdoor perturbations on Blind-P are $82.46$/$75.09$/$100.00$/$100.00\%$.}
    \label{fig:vis_blind_p_diverse}
\end{figure}

\paragraph{Diverse Initial Conditions} 
One reason for the good synthesis results in Figure~\ref{fig:vis_exp_blind} could be that the clean image already has a smooth background in the region of interest, i.e., top left corner, which could make the synthesis process easier. We investigate how SmoothInv is affected by the initialization of starting images. Note that for TrojAI result in Figure~\ref{fig:vis_exp_trojai_htba}, we already show one example with rainy effects and darker background where SmoothInv still synthesizes faithful backdoor patterns. Here we analyze the Blind-P model. We select two images from ImageNet testset with non-uniform color regions (high variance in pixel values) in the top-left corner as starting images. We also use two artificial inputs: images with pure white/black pixels. We show the results of SmoothInv on these images in Figure~\ref{fig:vis_blind_p_diverse}. We can see that with various conditions of starting images, our approach consistently synthesizes backdoor patterns in the top left region, while the synthesized perturbations itself achieve high ASRs as well.

Additional visualization results are provided in Appendix~\ref{appendix:visualization}, where we include a direct comparison between SmoothInv \textit{w/ diffusion} and \textit{w/o diffusion} in Figure~\ref{appendix:fig:diffusion_ablation}.

\subsection{Mitigation of Adaptive Attacks}
So far our experiments have focused on backdoor inversion on backdoored classifiers obtained from previous backdoor attacks. However, given our proposed method, someone could design new backdoor attacks to bypass our method, i.e. making it hard to extract effective backdoors with SmoothInv. The core step of SmoothInv is converting a standard backdoored classifier to a robust smoothed classifier, from which we can obtain perceptually-aligned gradients to reveal backdoor patterns. An adaptive attacker would try to circumvent SmoothInv by targeting the smoothing procedure: making the original backdoor ineffective for the smoothed classifier. Here we propose two adaptive attack attempts and show that SmoothInv is still robust in those challenging settings.

\begin{figure}[t!]
\centering 
        \includegraphics[width=1.0\linewidth]{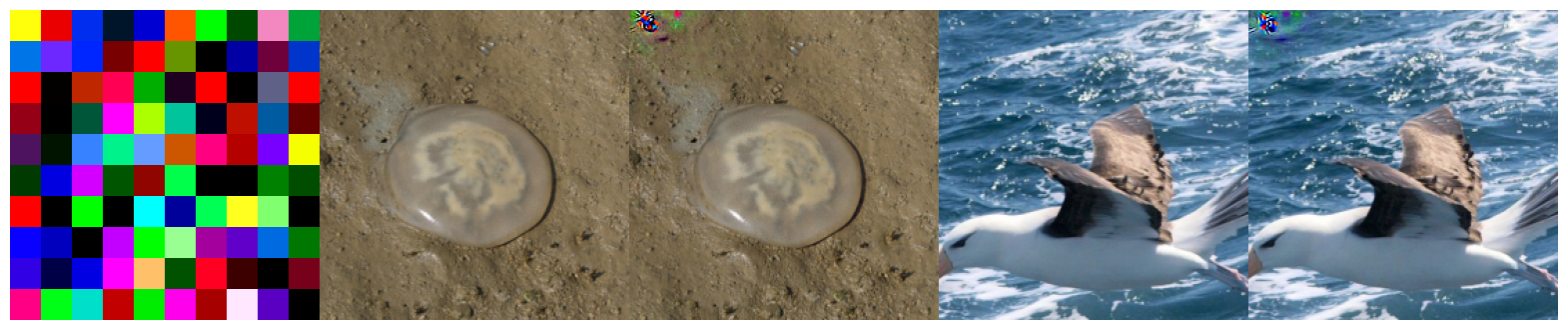}
    \caption{Results of SmoothInv \textit{w/o diffusion} on the backdoored classifier Blind-G with a \textit{gaussian backdoor} (leftmost). We show two pairs of clean and synthesized backdoored images ($\epsilon=10$). The two synthesized backdoor perturbations have an ASR of $88.78\%$ and $86.31\%$ respectively.}
    \label{fig:vis_exp_blind_g}
\end{figure}

\paragraph{Gaussian Backdoor} 
One can target the SmoothInv procedure by designing a backdoor which is hardly effective on the backdoored classifier after going through the smoothing process, i.e., $\mathcal{T}(\mathcal{B}(x)+\delta)$. One immediate choice is to use a backdoor with pure Gaussian noise: a gaussian backdoor $\mathcal{B}_{g}$. With such backdoor, the backdoor information can be obfuscated after the process $\mathcal{T}(\mathcal{B}_{g}(x)+\delta)$ as $\delta$ is also gaussian noise. 
We construct a \textit{gaussian backdoor} of size $10\times 10$, sampled from $\mathcal{N}(0,I)$ (see Figure~\ref{fig:vis_exp_blind_g} left). We use blind backdoor~\cite{bagdasaryan2020blind} to obtain a  backdoored ImageNet classifier with this gaussian backdoor, which we call Blind-G. We are able to achieve an ASR of $100.00\%$. 

On first inspection, we find that this simple \textit{gaussian backdoor} does invade the smoothing procedure of SmoothInv \textit{w/ diffusion}: 
the gaussian backdoor has an ASR of zero even for smoothed classifier constructed with noise level $\sigma=0.12$. We attribute this to the use of diffusion denoiser $\mathcal{D}$, where the Blind-G model becomes insensitive to the diffusion denoised backdoored images $\mathcal{D}(\mathcal{B}_{g}(x)+\delta)$. However, we find that the gaussian backdoor still remains highly effective for smoothed classifiers (SmoothInv \textit{w/ diffusion}) constructed purely from the Blind-G model (Equation~\ref{eq:def_smoothing}), despite a high drop of clean accuracy. We then apply SmoothInv \textit{w/o diffusion} to the Blind-G model and achieve an average ASR of $64.84\%/91.24\%$ ($\epsilon=10/20$) for reversed backdoors from single images. We visualize some synthesized backdoor patterns in Figure~\ref{fig:vis_exp_blind_g}. We can see that a dense colorful pattern emerges in the top left region via our SmoothInv process.

\paragraph{Training-Time Intervention} One  could also make the backdoor ineffective for the smoothed classifier by modifying the training procedure. For SmoothInv, the backdoored classifier sees the processed images $\mathcal{T}(x+\delta)$ instead of the original image $x$. An adaptive attacker can intentionally make the backdoored classifier misclassify backdoored images $\mathcal{B}(x)$ while classifying $\mathcal{T}(\mathcal{B}(x)+\delta)$ correctly. To investigate if this is possible, we design a new training objective below: (we consider SmoothInv \textit{w/o diffusion} due to resource limitations.)
\begin{equation}\label{eq:new_loss}
  \alpha_{0}\mathcal{L}(x,y)+\alpha_{1}\mathcal{L}(\mathcal{B}(x),y_{t})+\alpha_{2}\mathcal{L}(\mathcal{T}(\mathcal{B}(x)+\delta),y)
\end{equation}
We use the pixel pattern and single pixel backdoors in Figure~\ref{fig:stats_backdoor} and train classifiers with the new objective: Blind-P*/S* ($\alpha_{0},\alpha_{1},\alpha_{2}$ are automatically adjusted following~\cite{bagdasaryan2020blind}). We summarize the clean and backdoor accuracy of these models in Table~\ref{table:train_intervention}. We can see that the base classifiers Blind-P*/S* have lower backdoor ASR compared to Blind-P/S, suggesting that correctly classifying $\mathcal{T}(\mathcal{B}(x)+\delta)$ affects the effectiveness of backdoor attacks in a negative way. We also study how training-time intervention affects the effectiveness of SmoothInv on Blind-P* (attack considered successful). We find that we are still able to synthesize effective backdoor perturbations with an average ASR of $88.81\%$ over 10 random starting images. 

\begin{table}[t!]
\centering 
\Large 
\resizebox{1.00\columnwidth}{!}{
\begin{tabular}{lccc}
\Xhline{2pt}
         & \multicolumn{2}{c}{Base Classifier} &  Smoothed ($\sigma=0.25$) \\
         & Clean Acc   & Backdoor ASR   & Backdoor ASR   
          \\ \hline 
Blind-P  & 69.26$\%$          & 99.29$\%$      & 94.90$\%$                              \\
Blind-P* & 67.60$\%$           & 92.60$\%$     & 59.60$\%$                            \\ \hline 
Blind-S  & 68.06$\%$          & 79.73$\%$     & 59.20$\%$                            \\
Blind-S* & 66.60$\%$         & 45.80$\%$      & 31.40$\%$   \\
\Xhline{2pt}
\end{tabular}
}
\caption{Effect of training-time intervention on backdoor attacks.}
\label{table:train_intervention}
\end{table}

\section{Conclusion and Discussion}
In this paper, we have presented a method for backdoor inversion using a \emph{single} clean image from the underlying data distribution.  Unlike previous optimization-based approaches, our method exploits recent advances in adversarial robustness to create a smoothed version of a classifier, and then modify the image to extract the backdoor via this robust smoothed classifier.  We show that SmoothInv is able to recover backdoor perturbations that are both highly successful \emph{and} extremely visually similar to the true underlying backdoor.  Going forward, the work suggests that many current approaches to producing backdoored classifiers can easily be ``reverse engineered'' to recover the underlying backdoor, which can provide a powerful mechanism to analyze the security of existing classifiers.

One major limitation of SmoothInv is that it does not generalize to more advanced backdoors. In this work, we consider patch-based backdoor in particular. However, other forms of backdoor are shown possible by previous work, e.g. image wrapping~\cite{wanet2021nguyen}, adaptive imperceptible perturbation~\cite{Zhao_2022_CVPR} and instagram filters~\cite{darpa2021trojai}. Our method does not apply to those backdoors and we believe the likely reason is that $\ell_{2}$ based perturbations are only suitable for reversing patch-based backdoors. For advanced backdoors, we suspect that one would need to design the perturbation space of SmoothInv more carefully. For instance, we can model instagram filters with a per-pixel position-dependent transformation implemented by a neural network~\cite{hu2022topo}. It would be interesting future work to extend our approach beyond patch based backdoors.

\paragraph{Acknowledgments.} We thank Florian Tram\`er for valuable discussions during the development of this work. We thank Giulia Fanti, Aditi Raghunathan, Leslie Rice and Stephanie Rosenthal
 for reviewing prior draft of this work.  Mingjie Sun was supported by funding from the Bosch Center for Artificial Intelligence.

{\small
\bibliographystyle{ieee_fullname}
\bibliography{main}
}


\appendix

\begin{figure}[htbp]
\centering 
        \begin{minipage}[]{0.5\linewidth}
        {\includegraphics[width=1.0\linewidth]{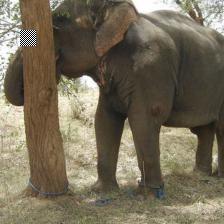}}
        \end{minipage}
    \caption{An image with a $16\times16$ backdoor used in~\cite{carlini2022clip}.}
    \label{appendix:fig:clip_backdoored_example}
\end{figure}

\begin{figure}[htbp]
\centering 
        \begin{minipage}[]{0.9\linewidth}
        {\includegraphics[width=1.0\linewidth]{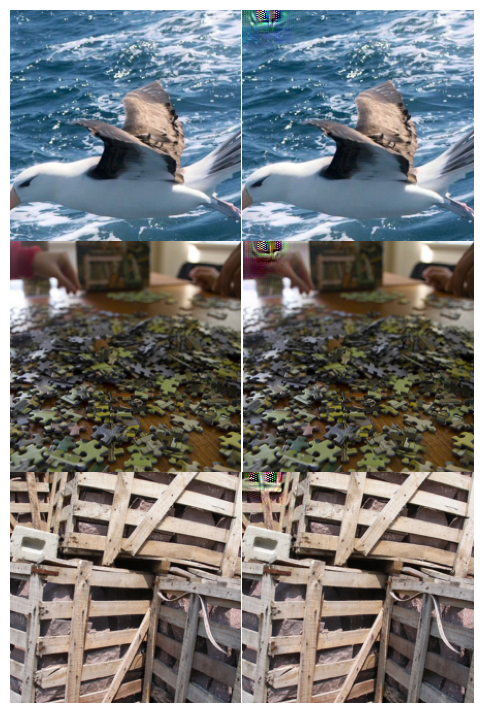}}
        \end{minipage}
    \caption{SmoothInv on a backdoored CLIP model.}
    \label{appendix:fig:clip_backdoor}
\end{figure}

\section{CLIP Backdoor~\cite{carlini2022clip}}~\label{appendix:clip_backdoor}
CLIP Backdoor~\cite{carlini2022clip} demonstrated a backdoor attack on a multimodal contrastive model CLIP~\cite{radford2021clip}. It uses a patch backdoor: a $16\times16$ grid of black and white pixels. We train a backdoored CLIP following the steps described in~\cite{carlini2022clip}. We visualize a sample backdoored image in Figure~\ref{appendix:fig:clip_backdoored_example}. The  backdoored CLIP model we trained has a backdoor ASR of $99.99\%$. We then cast this CLIP model as a standard image classifier in a zero-shot manner and apply SmoothInv \textit{w/o diffusion} only (due to limited computation resources). We test this on 10 random clean images from ImageNet testset and the average ASRs of the reversed backdoors are $73.16\%$ for ($\epsilon=5$) and $93.51\%$ for ($\epsilon=10$). We show the reversed backdoored images in Figure~\ref{appendix:fig:clip_backdoor}. We can see that some patterns with high color contrast appear in the top-left part.

\section{Runtime and Resource Considerations}~\label{appendix:resource}
 The major time bottleneck of our approach SmoothInv comes from back-propagating through the smoothed classifier $G_{b}$. SmoothInv \textit{w/o diffusion} is generally fast as diffusion model is not involved. Here we mainly study the resource consumption of SmoothInv \textit{w/ diffusion}. First we would like to clarify that for SmoothInv \textit{w/ diffusion}, we are not using the full standard reverse diffusion process but the one-shot denoising approach proposed in~\cite{carlini2022free}, where we apply only one diffusion step to obtain an estimate of the denoised image. All our experiments were run on four RTX A6000 GPU machines with 48685MiB GPU memory each. In Table~\ref{appendix:tab_time}, we report the time to synthesize an image using SmoothInv with various number of noise vectors $N$. We compare with the baseline ``Standard'' where we back-propagate through the standard base classifier (ResNet-18). We can see that the time spent scales linearly with the number of noisy vectors. In our experiments, we find that $N=10$ noise vectors are usually enough for a stable synthesis result. In this case, SmoothInv takes roughly around 5 mins to synthesize one backdoored image in one GPU. It would be interesting future work to investigate methods to speedup our synthesis process when a diffusion denoiser is used.
\begin{table}[t!]
\Large
\renewcommand{\arraystretch}{1.1}
\addtolength{\tabcolsep}{-0.5pt}
\resizebox{1.00\columnwidth}{!}{
\begin{tabular}{ccccccc}
\Xhline{2.0pt}
         & Standard & $N=1$    & $N=5$    & $N=10$   & $N=20$   & $N=40$    \\
         \hline 
\#GPUs   & 1        & 1      & 1      & 1      & 2      & 4       \\
Time/sec & 9.00     & 103.04 & 344.61 & 339.82 & 691.79 & 1014.72 \\
\Xhline{2.0pt}
\end{tabular}
}
\caption{Time (in seconds) and resource taken to synthesize an image (400 PGD iterations) for SmoothInv \textit{w/ diffusion}. We report the results with different number of noisy samples $N$. ``Standard'' corresponds to the PlainAdv baseline.}
\label{appendix:tab_time}
\end{table}

\begin{algorithm}[ht!]
\caption{SmoothInv  (PyTorch-style)}
\label{appendix:pseudocode}
\begin{lstlisting}[style=Pytorch,escapeinside={(@}{@)}]
# model: the backdoored classifier
# diffusion: class-unconditional diffusion model
# x_orig: a single clean image from victim class 
# y_t: target class
# delta: perturbation vector (@$\delta$@) to be optimized
# sigma: noise level (@$\sigma$@) for RS procedure(@~\cite{cohen2019certified,carlini2022free}@)
# n: number of Monte Carlo noise samples
# eps, alpha, steps: PGD hyper-parameters 


def backdoor_inversion(f, diffusion, x_orig, y_t):
    for _ in range(steps):
        x = x_orig + delta
        x_n = x.repeat(n,1,1,1)
        x_noise = x_n + torch.randn_like(x_n) * sigma # add isotropic Gaussian noise

        x_denoised = diffusion.denoise(x_noise) # optional
        y_prob = model(x_denoised) 
        y_est = y_prob.mean(dim=0) # estimated output of the smoothed classifier

        loss = criterion(y_est, y_t)
        loss.backward()

        delta += alpha * l2_normalize(delta.grad)
        delta = project(delta, eps)
        delta.grad.zero()

    return x_orig + delta
\end{lstlisting}
\end{algorithm}

\section{Pseudo-code}\label{appendix:pseudo-code}
We provide a PyTorch-style pseudo-code for our approach SmoothInv in~\cref{appendix:pseudocode}, where we apply projected gradient descent~\cite{pgd_madry} to synthesize backdoored patterns given a single image. We use a pre-trained diffusion model to build a robust smoothed classifier, following~\cite{carlini2022free}.

\section{Additional Visualization Results}~\label{appendix:visualization}
\begin{figure*}[htbp]
\centering 
        \begin{minipage}{1.00\linewidth}
        {\includegraphics[width=1.0\linewidth]{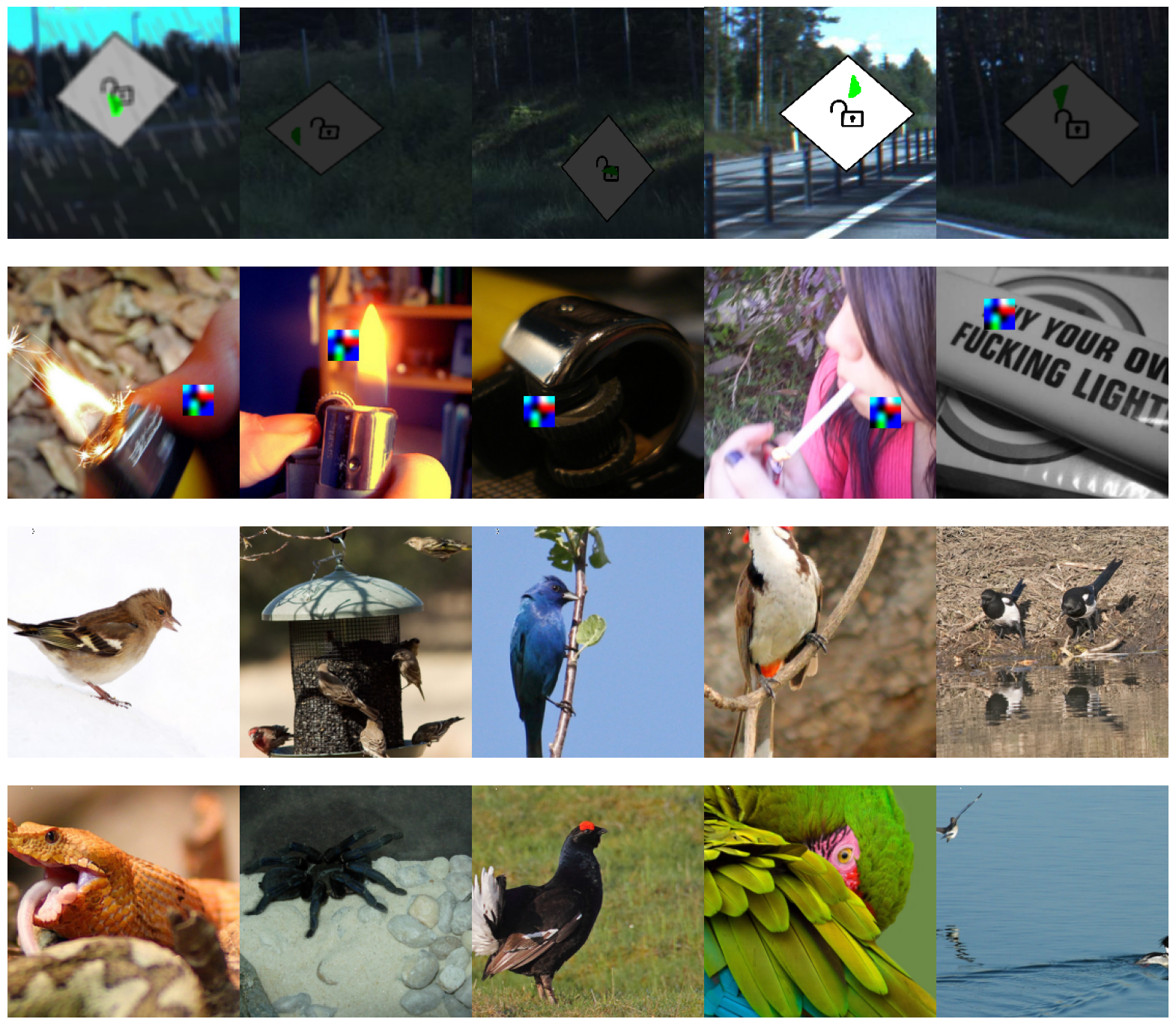}}
        \end{minipage}
        \vspace{-2ex}
    \caption{Backdoored images, from top to bottom: polygon/TrojAI, square/HTBA, pattern backdoor/Blind-P and single pixel/Blind-S.}
    \label{appendix:fig:vis_backdoored}
\end{figure*}

In Figure~\ref{appendix:fig:vis_backdoored}, we show some backdoored images with the true backdoors listed in Table~\ref{fig:stats_backdoor}. Notice that for Blind-P and Blind-S, the backdoors are placed in the top left region of the images (the injected backdoor may be hard to identify unless zooming in the specific part).

\begin{figure*}[ht]
\centering 
        \begin{minipage}[]{0.495\linewidth}
        \subcaptionbox{TrojAI-R4-131}
        {\includegraphics[width=1.0\linewidth]{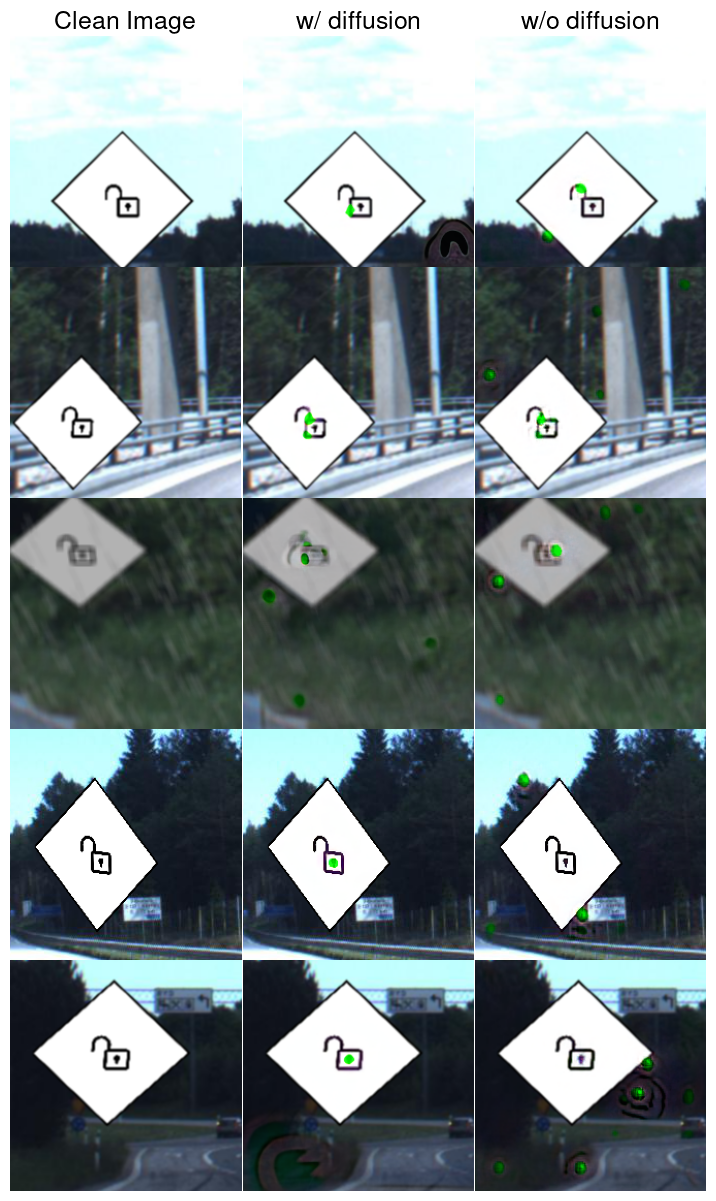}}
        \end{minipage} 
        \begin{minipage}[]{0.495\linewidth}
        \subcaptionbox{HTBA}
        {\includegraphics[width=1.0\linewidth]{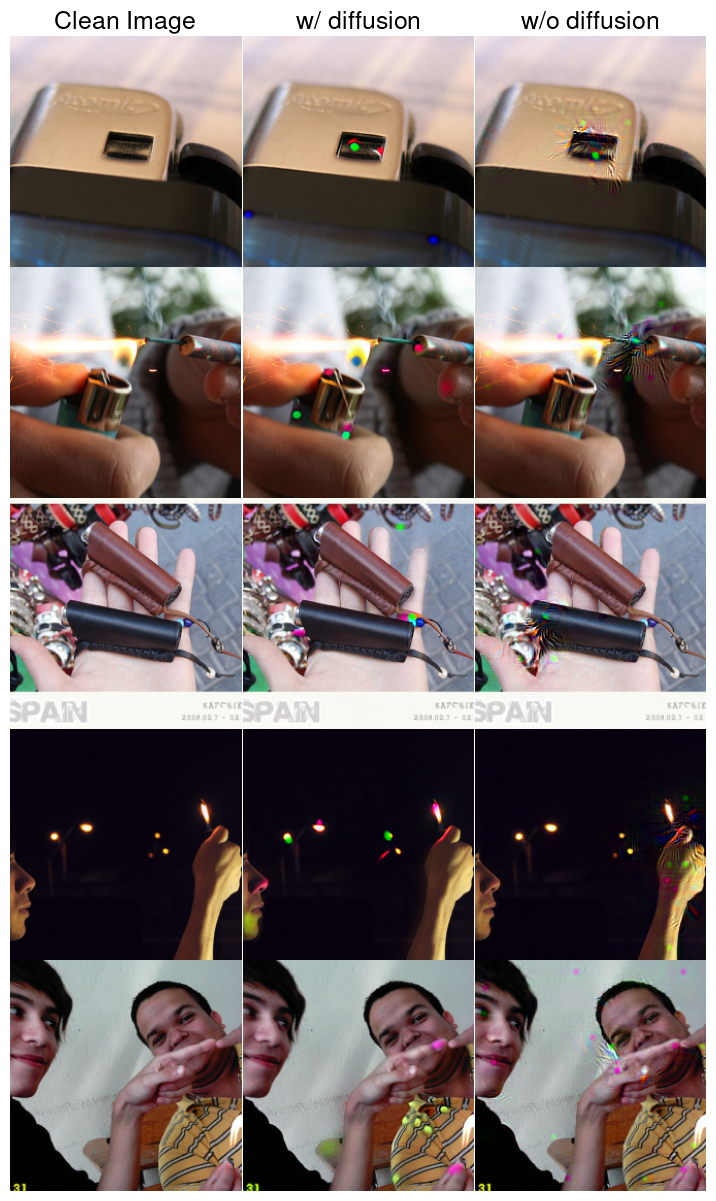}}
        \end{minipage}
    \caption{Comparison of SmoothInv \textit{w/ diffusion} and \textit{w/o diffusion} ($\epsilon=10$).}
    \label{appendix:fig:diffusion_ablation}
\end{figure*}

We provide a comparison of SmoothInv \textit{w/ diffusion} and \textit{w/o diffusion} in Figure~\ref{appendix:fig:diffusion_ablation}. We can see that for TrojAI, SmoothInv \textit{w/o diffusion} tends to generate more regions of interest on the background while the synthesized patterns appear more often in the foreground for SmoothInv \textit{w/ diffusion}. For HTBA, SmoothInv \textit{w/o diffusion} tends to have vague artifacts while the backdoor patterns are more distinctive for SmoothInv \textit{w/ diffusion}. This suggests that while SmoothInv \textit{w/o diffusion} may generate more effective backdoors, but using a diffusion denoiser may lead to better visualization results.

\begin{figure*}[htbp]
\centering 
        \begin{minipage}{1.00\linewidth}
        {\includegraphics[width=1.0\linewidth]{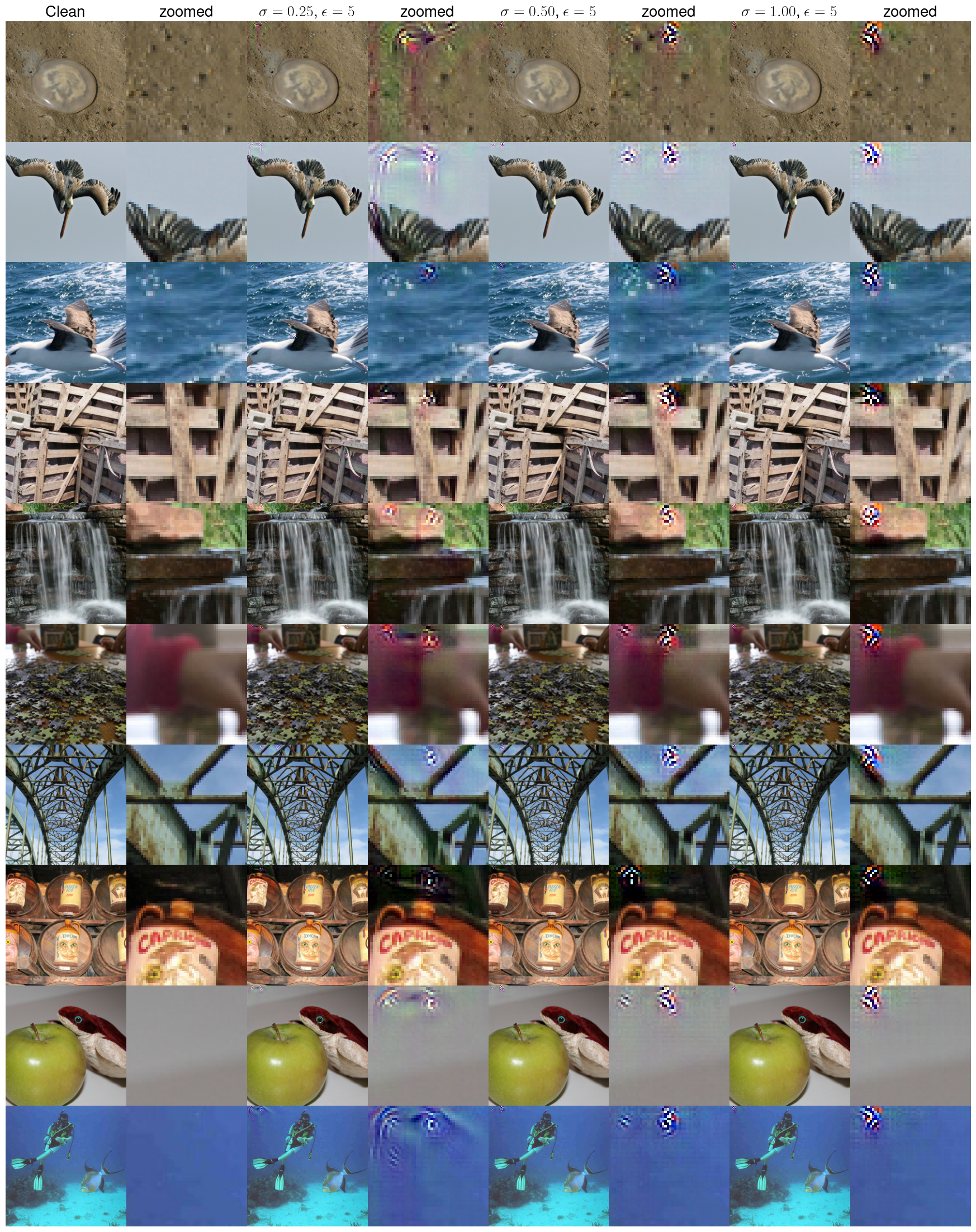}}
        \end{minipage}
        \vspace{-2ex}
    \caption{Additional results on the  Blind-P model.}
    \label{appendix:fig:vis_blind_p}
\end{figure*}

\begin{figure*}[htbp]
\centering 
        \begin{minipage}{1.00\linewidth}
        {\includegraphics[width=1.0\linewidth]{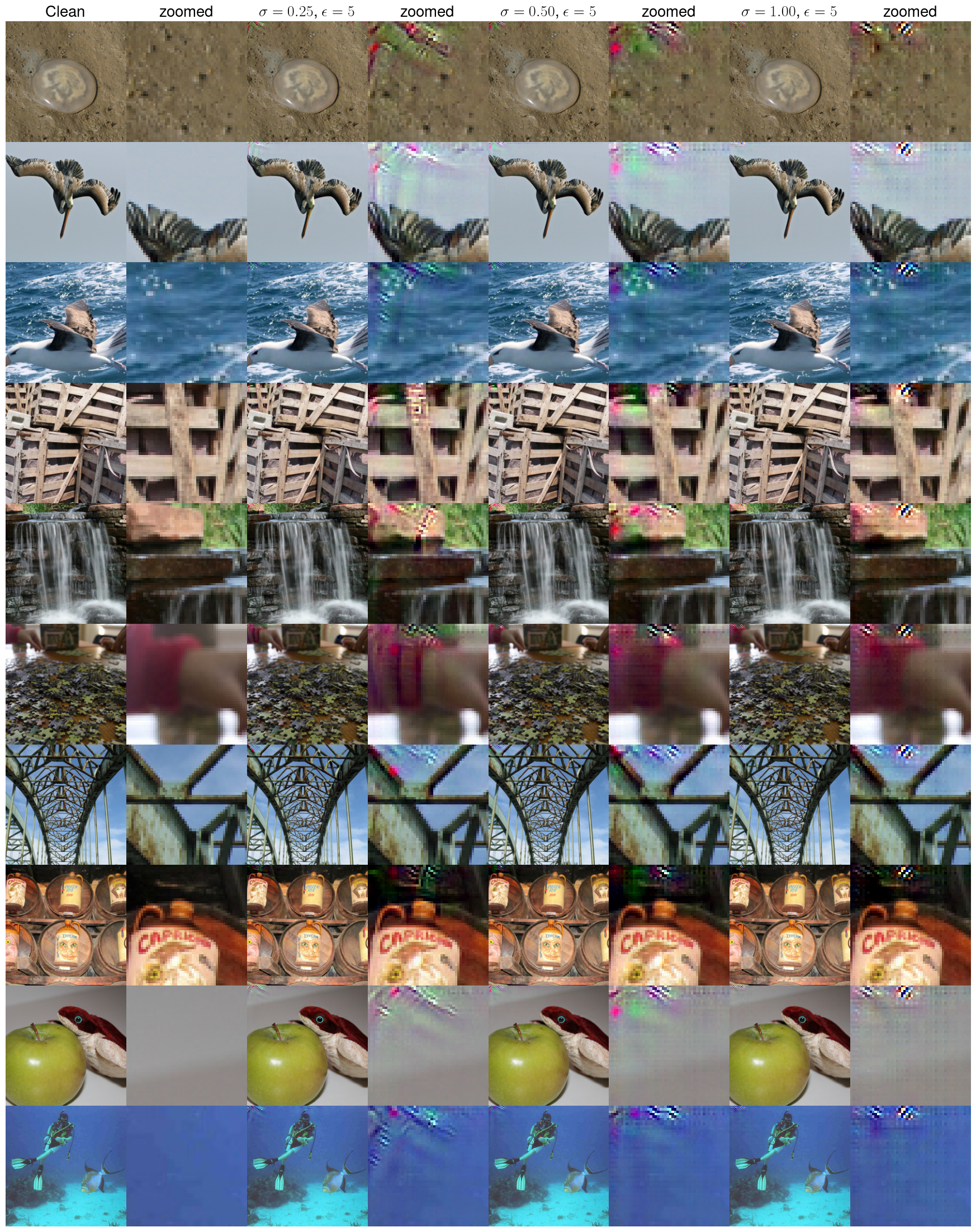}}
        \end{minipage}
        \vspace{-2ex}
    \caption{Additional results on the  Blind-S model.}
    \label{appendix:fig:vis_blind_s}
\end{figure*}

We include more visualization results ($\epsilon=10$) on the Blind-P and Blind-S models in Figure~\ref{appendix:fig:vis_blind_p} and Figure~\ref{appendix:fig:vis_blind_s}, where we show the synthesized images under various noise level $\sigma\in\{0.25, 0.50, 1.00\}$. We find no distinction between \textit{w/ diffusion} and \textit{w/o diffusion} visually so here we show the results of SmoothInv \textit{w/o diffusion} for these two models. 
We can see that in general, smoothed classifiers constructed with larger noise levels tend to give better visualization results.

\section{Sanity Check}~\label{appendix:sanity_check}
Following the initial sanity check result in Figure~\ref{fig:acc_blind_p}, we report the results on all four backdoored classifiers in Table~\ref{appendix:tab:sanity_check}. We show both the clean accuracy and backdoor ASR of the smoothed classifiers \textit{w/ } and \textit{w/o diffusion}. 
We can see that using a diffusion denoiser can significantly improve the clean accuracy of the resulting smoothed classifiers for all four backdoored classifiers. For backdoor ASR, we can see that the backdoor remains effective for smoothed classifiers both \textit{w/} and \textit{w/o diffusion} for some values of noise level $\sigma$.  
Note that for TrojAI, the results should have large variance as only five clean images are provided by the data publisher.

\newpage

\begin{table*}[ht!]
\centering 
\begin{minipage}[]{0.60\linewidth}
\Large
\renewcommand{\arraystretch}{1.1}
\addtolength{\tabcolsep}{-0.5pt}
\resizebox{1.00\columnwidth}{!}{
\begin{tabular}{ccrrrrr}
\Xhline{2.0pt}
    Model  &  Diffusion & 0.00 & 0.12 & 0.25 & 0.50 & 1.00 \\
    \hline
    \multirow{2}{*}{TrojAI} & \xmark & 100.00$\%$ & 100.00$\%$ & 80.00$\%$ & 20.00$\%$  & 0.00$\%$  \\
    & \cmark & 100.00$\%$ & 100.00$\%$ & 100.00$\%$ & 100.00$\%$ & 100.00$\%$\\ \hline 
    \multirow{2}{*}{HTBA} & \xmark & 95.00$\%$ & 2.00$\%$ & 0.00$\%$ & 0.00$\%$ & 0.00$\%$\\ 
    & \cmark & 95.00$\%$ &  90.00$\%$ & 92.00$\%$ & 98.00$\%$ & 96.00$\%$ \\ \hline 
    \multirow{2}{*}{Blind-P} & \xmark & 69.26$\%$ & 33.40$\%$ & 2.80$\%$ & 0.00$\%$ & 0.00$\%$\\
    &\cmark & 69.26$\%$ & 66.10$\%$  & 63.10$\%$  & 57.70$\%$  & 47.10$\%$  \\
    \hline 
    \multirow{2}{*}{Blind-S} &\xmark & 68.06$\%$ & 29.90$\%$ & 2.30$\%$ & 0.10$\%$ & 0.10$\%$ \\
    &\cmark  & 68.06$\%$ & 65.60$\%$ & 62.30$\%$ & 56.70$\%$ & 47.80$\%$\\
\Xhline{2.0pt}
\end{tabular}
\label{appendix:sanity_check_backdoor}
}
\subcaption{Clean Accuracy}
\end{minipage}

\vspace{2ex}
\begin{minipage}[]{0.60\linewidth}
\Large
\renewcommand{\arraystretch}{1.1}
\addtolength{\tabcolsep}{-0.5pt}
\resizebox{1.00\columnwidth}{!}{
\begin{tabular}{ccrrrrr}
\Xhline{2.0pt}
    Model  &  Diffusion & 0.00 & 0.12 & 0.25 & 0.50 & 1.00 \\
    \hline
    \multirow{2}{*}{TrojAI} & \xmark & 100.00$\%$ & 100.00$\%$ & 40.00$\%$ & 40.00$\%$ & 20.00$\%$ \\
    & \cmark & 100.00$\%$ & 100.00$\%$ & 100.00$\%$ & 60.00$\%$ & 40.00$\%$ \\ \hline 
    \multirow{2}{*}{HTBA} & \xmark & 54.00$\%$ &  70.00$\%$ & 100.00$\%$ & 100.00$\%$ & 100.00$\%$\\ 
    & \cmark & 54.00$\%$ & 64.00$\%$ & 58.00$\%$ & 64.00$\%$ & 48.00$\%$\\ \hline 
    \multirow{2}{*}{Blind-P} & \xmark & 99.29$\%$ & 99.80$\%$ & 94.90$\%$& 40.10$\%$ & 4.70$\%$ \\
    &\cmark & 99.29$\%$ & 99.40$\%$ & 87.70$\%$ & 1.70$\%$ & 0.10$\%$ \\
    \hline 
    \multirow{2}{*}{Blind-S} &\xmark & 79.73$\%$ & 89.60$\%$ & 88.40 $\%$ & 81.70$\%$ & 97.00$\%$\\
    &\cmark  & 79.73$\%$ & 59.20$\%$ & 21.50$\%$  & 4.00$\%$ & 0.00$\%$ \\
\Xhline{2.0pt}
\end{tabular}
\label{appendix:sanity_check_clean}
}
\subcaption{Backdoor ASR}
\end{minipage}
\caption{Clean accuracy and backdoor ASR of the smoothed classifiers (\textit{w/} and \textit{w/o diffusion}) with various values of $\sigma$: 0.12, 0.25, 0.50 and 1.00. The $\sigma=0$ column corresponds to the results of the base backdoored classifier. The results on TrojAI are computed on a limited number of $5$ available clean images so they should have high variances.
}
\label{appendix:tab:sanity_check}
\end{table*}

\end{document}